\documentclass[letterpaper, 10 pt, conference]{ieeeconf}

\usepackage{amsmath,amsfonts}
\usepackage{array}
\usepackage[caption=false,font=normalsize,labelfont=sf,textfont=sf]{subfig}
\usepackage{textcomp}
\usepackage{verbatim}
\usepackage{graphicx}
\usepackage{longtable}
\usepackage{booktabs}
\usepackage{multirow}
\usepackage[table]{xcolor}

\usepackage{algorithm}
\usepackage{algpseudocode}

\IEEEoverridecommandlockouts
\begin{document}

\title{\LARGE \bf \textit{TeTRA}-VPR: A Ternary Transformer Approach for Compact Visual Place Recognition}

\author{Oliver Grainge$^{1}$, Michael Milford$^{2}$, Indu Bodala$^{1}$, Sarvapali D. Ramchurn$^{1}$ and Shoaib Ehsan$^{1, 3}$%
\thanks{Manuscript received: -; Revised -; Accepted -}%
\thanks{This paper was recommended for publication by Editor - upon evaluation of the Associate Editor and Reviewers' comments.
This work was supported by the UK Engineering and Physical Sciences Research Council through grants EP/Y009800/1 and EP/V00784X/1}
\thanks{$^{1}$O. Grainge, I. Bodala. S. D. Ramchurn and S. Ehsan are with the school of Electronics and Computer Science, University of Southampton, United Kingdon {\tt\small (email: oeg1n18@soton.ac.uk; i.p.bodala@soton.ac.uk; sdr1@soton.ac.uk; s.ehsan@soton.ac.uk).}}%
\thanks{$^{2}$M. Milford is with the School of Electrical Engineering and Computer Science, Queensland University of Technology, Brisbane, QLD 4000, Australia {\tt\small (email: michael.milford@qut.edu.au).}}%
\thanks{$^{3}$S. Ehsan is also with the school of Computer Science and Electronic Engineering, University of Essex, United Kingdom, {\tt\small (email: sehsan@essex.ac.uk).}}%
\thanks{Digital Object Identifier (DOI): see top of this page.}
}


\maketitle
\thispagestyle{empty}

\begin{abstract}
Visual Place Recognition (VPR) localizes a query image by matching it against a database of geo-tagged reference images, making it essential for navigation and mapping in robotics. Although Vision Transformer (ViT) solutions deliver high accuracy, their large models often exceed the memory and compute budgets of resource-constrained platforms such as drones and mobile robots. To address this issue, we propose \textit{TeTRA}, a ternary transformer approach that progressively quantizes the ViT backbone to 2-bit precision and binarizes its final embedding layer, offering substantial reductions in model size and latency. A carefully designed progressive distillation strategy preserves the representational power of a full-precision teacher, allowing \textit{TeTRA} to retain or even surpass the accuracy of uncompressed convolutional counterparts, despite using fewer resources. Experiments on standard VPR benchmarks demonstrate that TeTRA reduces memory consumption by up to 69\% compared to efficient baselines, while lowering inference latency by 35\%, with either no loss or a slight improvement in recall@1. These gains enable high-accuracy VPR on power-constrained, memory-limited robotic platforms, making \textit{TeTRA} an appealing solution for real-world deployment.
\end{abstract}

\section{Introduction}
Visual Place Recognition (VPR) is essential in a broad range of robotics applications, including navigation, mapping, and long-term autonomy. By identifying the location of a query image through visual matching against a geo-tagged database, VPR provides a robust localization capability in environments where GPS or other external signals may be unreliable. Commonly, VPR is framed as an image retrieval problem, where the goal is to extract discriminative features from an input image and retrieve the most similar images from a large database using vector search \cite{geobench, vprsurvey, vprsurveydeep}. In this setting, the challenge lies in designing effective feature representations and similarity measures that can reliably match images despite variations in appearance, viewpoint, and other environmental conditions.

\begin{figure}[!t]
\centering
\includegraphics[width=\columnwidth]{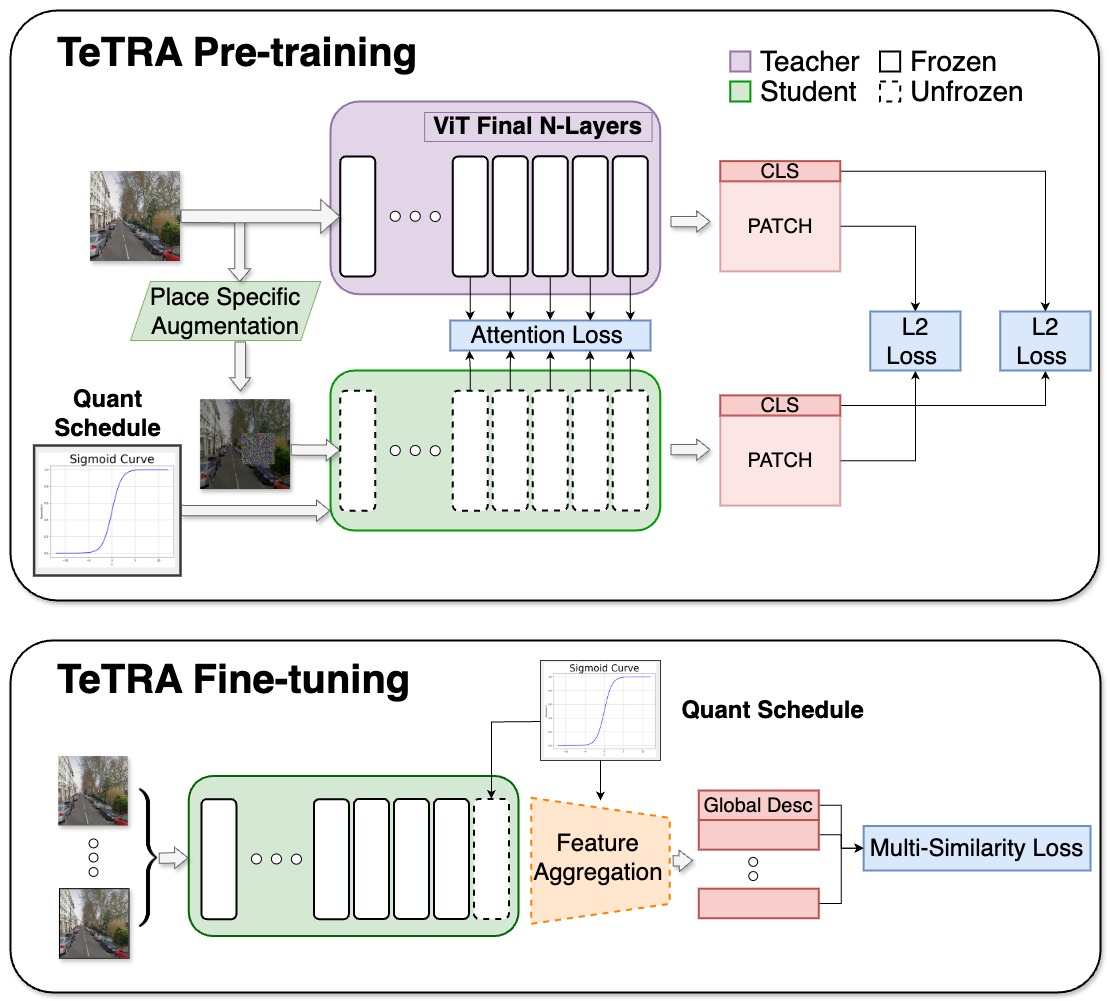}\label{fig:tetra}
\caption{\textit{TeTRA} block diagram illustrating the ternary and binary training pipeline. The pre-training stage employs distillation-based, progressive quantization-aware training with attention, classification, and patch token losses. During fine-tuning, all backbone layers except the last are frozen while supervised contrastive learning is performed using a multi-similarity loss.}
\label{fig:your_label}
\end{figure}

To achieve this objective, recent approaches have leveraged large scale vision foundation models such as DinoV2 to enhance VPR robustness and accuracy \cite{dinov2}. Many studies have demonstrated that these approaches maintain high accuracy and robustness against natural variations that occur in images of the same location, for instance changes in lighting, occlusions, viewpoint and appearance \cite{anyloc, boq, salad}. However, these methods typically incur significant computational and memory overhead \cite{quantized, distillation, pruning, binary}. This challenge is further compounded by the common use of multi-stage inference pipelines in which hierarchical coarse to fine retrieval mechanisms are employed, thereby increasing the cost of feature extraction, embedding and similarity computation \cite{r2former, selavpr}. Although smaller convolutional architectures can be used to reduce complexity, they often fall short of the accuracy achieved by methods based on foundation ViTs, particularly in scenarios with significant appearance changes \cite{boq, eigenplaces}.

In order to reduce the computational cost of applying foundation Vision Transformers (ViTs), inference optimization techniques can be employed. These include distillation, which compresses the knowledge of a large teacher model into a smaller student model; pruning, which removes redundancies by eliminating unnecessary weights, groups, or structures; and quantization, which lowers the precision of weights and activations to reduce memory requirements and alleviate bandwidth bottlenecks, ultimately accelerating inference \cite{distillation, pruning, quantized}.

Despite the success of transformer models in VPR, distillation and pruning have primarily been applied to convolution-based models, with quantization following a similar trend \cite{distillation, pruning, quantized}. In particular, 8-bit integer quantization has been shown to significantly reduce resource usage with minimal impact on accuracy \cite{quantized}. However, in the extreme quantization case where weights and activations are fully binarized, the efficiency gains come at the cost of substantial accuracy degradation \cite{xnornet, bnnsurvey}. Recently, in natural language processing tasks, ternary networks that use just one additional bit per weight compared to binary networks have demonstrated near full precision accuracy while achieving up to a sixteen fold reduction in memory usage \cite{bitnet}.

Building upon these insights, we introduce \textit{TeTRA}, a two stage training pipeline that synergizes distillation with ternary quantization of a transformer-based architecture, guided by a progressive quantization schedule. In \textit{TeTRA}, knowledge transfer is carefully structured to preserve the representations of the full precision teacher while aggressively reducing bit widths of the student. The backbone is quantized to ternary precision, while embeddings are further compressed using binary representations. Given the inherent challenges of such extreme quantization, particularly with regard to convergence stability and performance degradation \cite{xnornet, bnnsurvey}, we propose a progressive distillation strategy that incrementally refines both ternary and binary representations smoothing the transition from full precision training to quantized aware training.

Extensive experiments demonstrate that \textit{TeTRA} achieves retrieval accuracy on par with full-precision convolution models while significantly reducing model size and inference latency. The main contributions of this work are:

\begin{itemize}
    \item \textbf{Ultra-Low-Bit Quantization for Transformers in VPR:} 
    We propose \textit{TeTRA}, a method that compresses a Vision Transformer by quantizing its backbone weights to ternary and its final embedding layer to binary, while preserving the expressive power necessary for robust place recognition under severe appearance changes.

    \item \textbf{Progressive Distillation for Stable Extreme Quantization:} 
    We develop a progressive distillation pipeline that gradually transitions from full-precision to extremely low-bit training. This pipeline leverages multi-level supervision, aligning classification tokens, patch tokens, and attention maps with a powerful teacher model to maintain high-quality representations.

    \item \textbf{State-of-the-Art Memory Efficiency and Pareto-Optimal Performance:} 
    We show that \textit{TeTRA} achieves up to 69\% lower model size and 35\% lower latency compared to standard baselines, while maintaining or improving recall across diverse VPR benchmarks. This delivers a new Pareto frontier of memory and latency versus accuracy for resource-constrained robotic applications.
\end{itemize}

This paper is structured as follows: Section \ref{sec:related_work} reviews related work on visual place recognition and efficient neural networks. Section \ref{sec:method} details our \textit{TeTRA} training pipeline, including pre-training and fine-tuning. Section \ref{sec:results} presents experiments on VPR benchmarks, evaluating accuracy, memory, and latency. Finally, Section \ref{sec:conclusion} discusses key findings and future directions.

\section{Related Work}
\label{sec:related_work}

VPR has been a longstanding research topic in robotics and computer vision, with applications spanning localization, navigation, and long-term autonomy. Early methods relied on hand-crafted features, whereas modern approaches leverage deep learning to learn more robust and discriminative representations.

\subsection{Early VPR Methods}
Initial VPR feature extraction techniques were based on local keypoint descriptors such as SIFT, SURF, and RootSIFT \cite{sift,surf,rootsift}. These features were typically aggregated into global representations using methods like Bag-of-Words (BoW) and VLAD, enabling efficient image retrieval \cite{bow,vlad}. While computationally efficient, their reliance on manually designed descriptors made them vulnerable to changes in viewpoint, illumination, and appearance \cite{vprbench}.

\subsection{CNN-based Approaches}
The introduction of Convolutional Neural Networks (CNNs) significantly improved VPR by learning hierarchical, data-driven features. NetVLAD, for example, combined a CNN backbone with a learnable VLAD layer, often trained using contrastive learning, to generate powerful yet high-dimensional embeddings \cite{netvlad}. Classification-based methods like EigenPlaces and CosPlace further reduced descriptor dimensionality while preserving strong retrieval performance \cite{eigenplaces,cosplace}. Large-scale supervised fine-tuning, such as on GSV-Cities, further enhanced accuracy \cite{gsvcities}. However, CNN-based methods can still struggle with severe environmental changes, and their high-dimensional descriptors consume large amounts of database memory \cite{boq, netvlad}. 

\subsection{Transformer-based Approaches}
Inspired by the success of Transformers \cite{vaswani2017attention}, more recent works have explored ViTs for VPR. Methods such as SALAD, BoQ, and AnyLoc utilize the very general visual representations of large-scale, self-supervised ViT backbones, often with parameter-efficient fine-tuning, to improve robustness against extreme viewpoint and appearance variations~\cite{salad,boq,anyloc}. However, these approaches typically require significant computational and memory resources, especially in multi-stage retrieval pipelines \cite{r2former,selavpr}.

\subsection{Efficiency Optimization Techniques}
To reduce model size and inference latency, efficiency-oriented strategies such as 8-bit integer quantization, structured network pruning, and knowledge distillation have been explored for CNN-based VPR, achieving compression with minimal accuracy loss~\cite{quantized,pruning,distillation}. More extreme compression techniques, such as binary quantization (e.g., FloppyNet), further reduce memory and computation requirements by replacing matrix multiplications with XOR and bitcount operations \cite{binary}. However, these methods often result in significant accuracy degradation due to their extreme constraints limiting representational capacity~\cite{binary}. While these techniques have been extensively studied for CNNs, their application to Transformers in VPR remains largely unexplored.

\subsection{Ternary Quantization}
As an alternative to binary quantization, ternary quantization provides a balance between extreme compression and representational capacity. By restricting model weights to $\{-1, 0, 1 \}$, it enables aggressive compression while replacing matrix multiplications with sparse additions and subtractions, significantly improving energy efficiency \cite{binary-ternary}. This optimization also facilitates the design of specialized accelerators \cite{ternmatmul,ternhardware}, which prioritize parallel adders over multipliers, further enhancing efficiency \cite{bitnet}. Unlike binary networks, which often suffer from severe accuracy degradation, ternary models have demonstrated near full-precision performance in NLP applications while offering substantial resource savings \cite{xnornet,bitnet}. However, despite these advantages, the potential of ultra-low-bit quantization for ViT-based VPR remains largely unexplored.

\section{Method}\label{sec:method}
In this section, we introduce the design and training strategies of \textit{TeTRA}, our distillation-based, progressive quantization-aware framework for efficient VPR. We begin by describing the modifications we make to the standard ViT architecture, followed by details of the pre-training process and, finally, the fine-tuning stage.

\subsection{Model Architecture}
\textit{TeTRA} is based on the standard ViT architecture, specifically the base configuration with 12 layers, 12 attention heads, and a hidden dimension of 768 \cite{vit16x16}. However, we introduce a few minor modifications. In line with previous work on low-bit-width quantized transformers, we incorporate additional normalization layers to maintain activation variance and reduce quantization noise \cite{bitnet}. Specifically, within the multi-head self-attention (MHSA) mechanism each head is defined as:

\begin{equation}\label{eqn:out_proj} \text{head}_h = \operatorname{Attention}\Bigl(\mathbf{X},\widetilde{\mathbf{W}}_h^Q, \mathbf{X},\widetilde{\mathbf{W}}_h^K, \mathbf{X},\widetilde{\mathbf{W}}_h^V\Bigr) \end{equation}

Where $X$, is in the input tokens and ($\widetilde{\mathbf{W}}_h^Q, \widetilde{\mathbf{W}}_h^K, \widetilde{\mathbf{W}}_h^V$) are the query, key and value projection matrices of head $h$.  After computing attention for each head, we concatenate the results and apply an additional layer normalization before performing the final output projection:

\begin{equation} \operatorname{MHSA}(\mathbf{X}) = \operatorname{LN}\Bigl( \operatorname{Concat}\bigl(\text{head}_1, \dots, \text{head}_H\bigr) \Bigr)\widetilde{\mathbf{W}}^O \end{equation}

Similarly, in the feedforward network (FFN), we introduce an additional layer normalization before the down-projection, as expressed by Equation \ref{eqn:ffn}. 

\begin{equation}\label{eqn:ffn} \operatorname{FFN}(\mathbf{X}) = \operatorname{LN}\Bigl( \operatorname{GELU}\bigl(\mathbf{X}\widetilde{\mathbf{W}}_1 + \mathbf{b}_1\bigr) \Bigr)\widetilde{\mathbf{W}}_2 + \mathbf{b}_2 \end{equation}

\subsection{Pre-training}
Ternary quantization fundamentally reduces the representational capacity of a model by constraining its weights to the set $\{-1, 0, 1\} $. Since full precision networks are not trained with this constraint, their weight parameters are often incompatible. Therefore, applying post-training quantization to a pre-trained model introduces significant errors that severely degrade accuracy, making direct conversion impractical \cite{bitnet, binary-ternary}. Consequently, ternary networks must be trained to be robust to these constraints, a goal typically achieved through quantization-aware training (QAT) \cite{bivit}. However, QAT introduces gradient estimation errors that can lead to poor convergence and reduced accuracy \cite{bnnsurvey, xnornet}. To mitigate this challenge while adhering to ternary quantization constraints, we employ a progressive quantization schedule that gradually transitions the model from full-precision training to fully quantized training with ternary weight constraints.

To further enhance the model's robustness and generalization despite these constraints, we use a strong and scalable supervision signal through knowledge distillation from a visual foundation model on unlabeled data. Additionally, during training, we apply targeted augmentation techniques exclusively to the student model to encourage the learning of useful invariances, optimizing \textit{TeTRA}'s effectiveness for the place recognition task.

\begin{figure*}[!t]
\centering
\includegraphics[width=\textwidth]{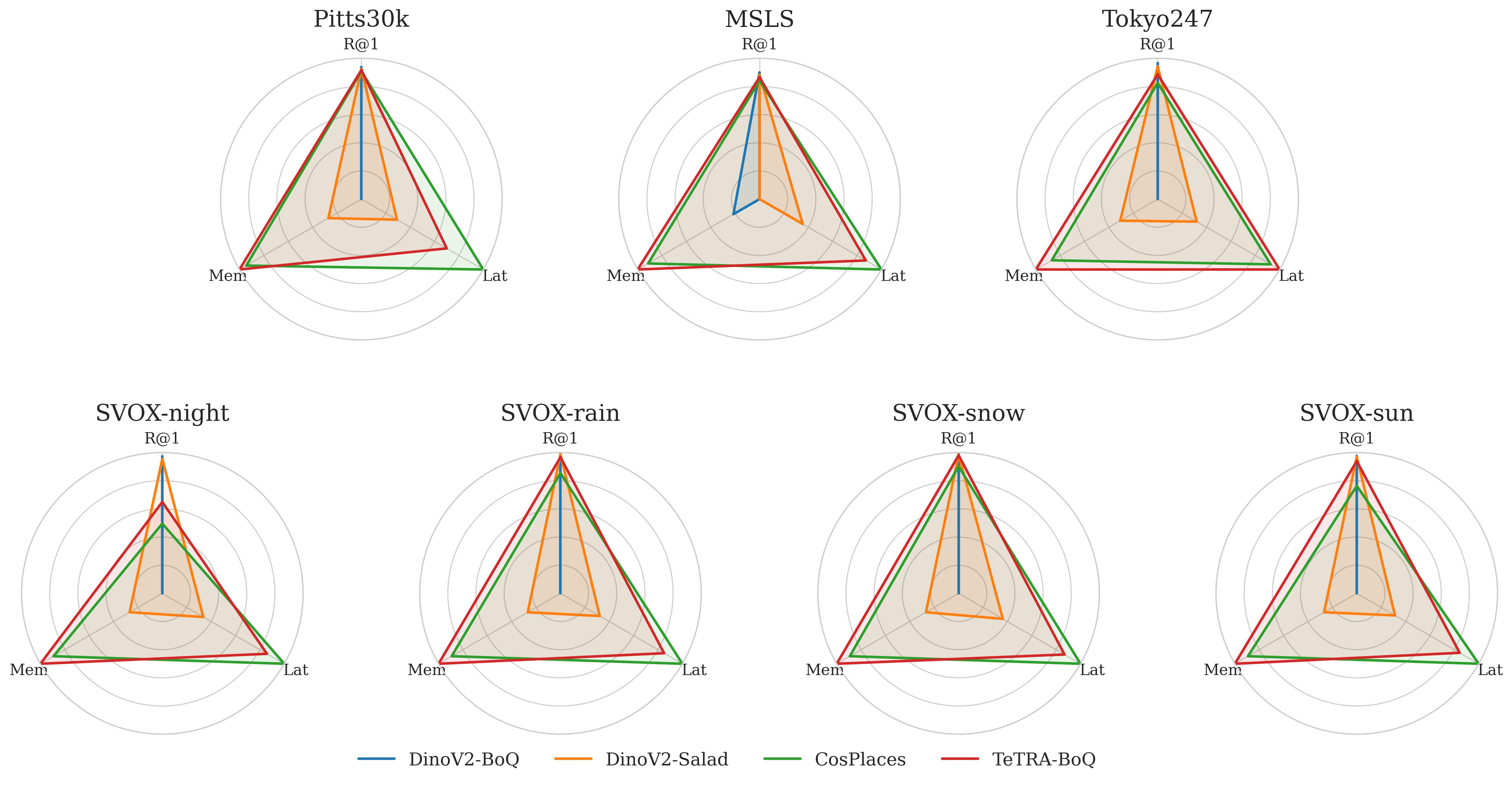}\label{fig:radar}
\caption{Radar plot of normalized metrics comparing inverse memory usage efficiency (Mem), matching speed (Lat), and R@1 accuracy across multiple datasets. Higher values indicate better performance for each metric. The results show that DinoV2-based models consume more resources, whereas \textit{TeTRA} achieves higher R@1 accuracy, especially on appearance change datasets while using less memory than CosPlace.}
\end{figure*}

\subsubsection{Quantization-Aware Training}
Our ternary quantization method utilizes Abs-Mean quantization to reduce the weight bit-width to 2, thereby decreasing static memory consumption by a factor of 16 and lowering memory bandwidth load during inference. Given a weight matrix 
$W$, the weight quantization function $Q_w$ is defined as:

\begin{equation}\label{eqn:w_quant}
    \widetilde{W} = Q_W(W) = \gamma\,\operatorname{clamp}\!\left(\frac{W}{\gamma + \epsilon},\, -1,\, 1\right)
\end{equation}

where $\epsilon$ is a small offset to prevent division by zero, and $ \gamma $ is the per-tensor scaling factor, computed as the mean absolute value of the weight:
 
\begin{equation}\label{eqn:gamma}
    \gamma = \frac{1}{MN}\sum_{i=1}^{M}\sum_{j=1}^{N}|W_{i,j}|
\end{equation}

\textit{TeTRA} uses per-tensor quantization granularity for weights with just a single scale parameter, $ \gamma$, for each weight matrix. However, activations ($X$), which exhibit greater channel-wise variation and dynamic range \cite{smoothquant}, benefit from finer quantization granularity and increased precision. To this end, \textit{TeTRA} employs 8-bit per-token activation quantization, using linear-symmetric scaling \cite{quantized}, with one scale parameter $s_j$ per-token of the activation tensor thereby minimizing information loss. This quantization procedure is given by equation \ref{eqn:a_quant}.

\begin{equation}\label{eqn:a_quant}
    \widetilde{X}_j = Q_x(X_j) = s_c\,\operatorname{round}\!\left(\frac{X_j}{s_j}\right)
\end{equation}

where \(s_j\) is the scale factor per channel of token $j$ in the sequence, that maps the full precision activation range to the quantized activation range. It is computed as followed, where \(b\) denotes quantized bit-width:

\begin{equation}\label{eqn:act_scale}
    s_j = \frac{\max |X_j|}{2^{b-1} - 1}
\end{equation}

Since \(Q_W\) and \(Q_A\) are non-differentiable due to the rounding operation, we adopt the conventional QAT method of using the Straight-Through Estimator (STE) as shown in equation \ref{eqn:ste} to pass gradients directly through the quantization functions \cite{bnnsurvey}. 

\begin{equation}\label{eqn:ste}
    \frac{\partial \widetilde{W}}{\partial W} \approx
    \begin{cases}
        1, & \text{if } |W| \le \gamma, \\[1mm]
        0, & \text{otherwise}
    \end{cases}
\end{equation}

However, to mitigate the gradient approximation errors caused by equation \ref{eqn:ste} and improve convergence, we introduce a progressive quantization scheduling parameter, $\lambda(t)$, which is parameterized by a sigmoid function of the training step $t$:

\begin{equation}\label{schedule_lambda}
    \lambda(t) = \frac{1}{1 + \exp\!\left(-\alpha\,t + \beta\right)}
\end{equation}

As $t$ increases, $\lambda(t)$ moves from $0$ to $1$, gradually shifting the network from using full-precision weights $W$ to their quantized counterparts $Q_w(W)$. The effective weights at step $t$ are defined by:

\begin{equation}\label{schedule_weights}
    \widetilde{W}(t) = \bigl(1 - \lambda(t)\bigr)\,W + \lambda(t)\,Q_w(W)
\end{equation}

During training, only the forward pass is quantized, while the backward pass updates the latent full-precision weights $W$ using gradients derived from the quantized forward pass. This progressive approach stabilizes training and improves convergence by reducing gradient estimtation error and avoiding an abrupt switch to quantized weights.

\begin{figure*}[!h]
\centering
\includegraphics[width=\textwidth]{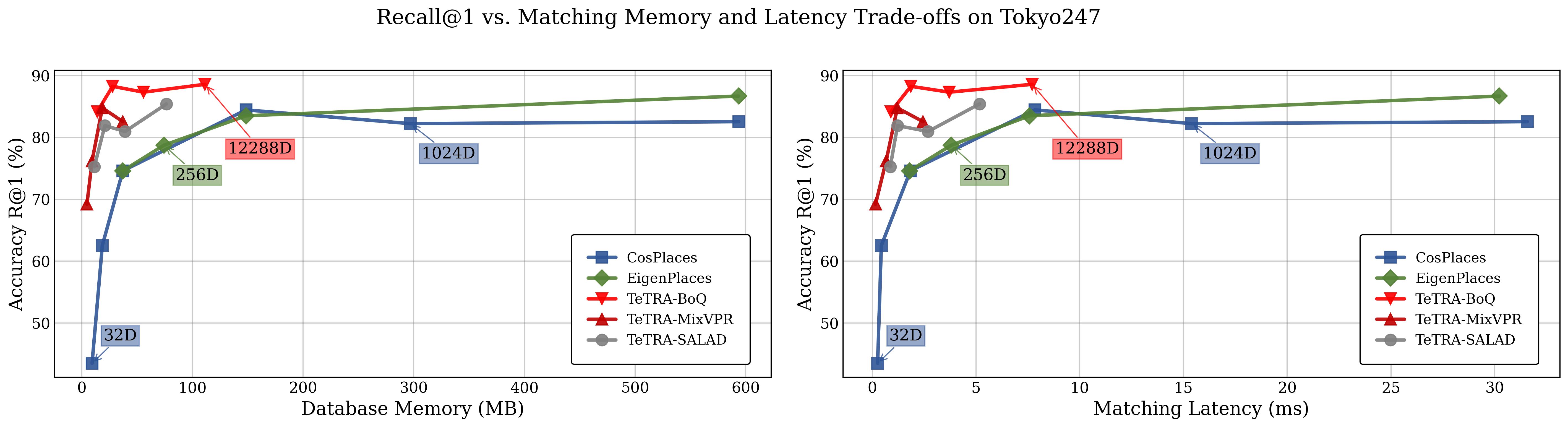}
\caption{Line plot demonstrating the trade-off between recall@1 accuracy and resource consumption (latency and memory usage) on the Tokyo247 dataset. Each line represents a single model assessed across different descriptor sizes, with selected points annotated to indicate the descriptor dimension (e.g., 1024D for 1024 dimensions).}
\label{fig:matching}
\end{figure*}

\subsubsection{Distillation}
For scalable training on unlabeled data, we use a learning signal from a full-precision teacher model. In particular, we employ the Bag of Learnable Queries (BoQ) ViT backbone, fine-tuned from the DinoV2 model \cite{dinov2}, as it achieves the highest Recall@1 accuracy among single-stage VPR feature extractors \cite{boq}.

As illustrated in Fig. \ref{fig:tetra}, our distillation approach incorporates three loss components, each scaled by its respective hyperparameter:

\begin{equation}
    \mathcal{L}_\text{cls} = \lambda_{\text{cls}} \sum_{i=1}^N \left\| T_\text{cls}(x; W_T)_i - S_\text{cls}\bigl(x'; \widetilde{W}_\text{S}(t)\bigr)_i \right\|_2^2
\end{equation}
\begin{equation}
    \mathcal{L}_\text{tok} = \lambda_{\text{tok}} \sum_{i=1}^N \left\| T_\text{tok}(x; W_T)_i - S_\text{tok}\bigl(x'; \widetilde{W}_\text{S}(t)\bigr)_i \right\|_2^2
\end{equation}
\begin{equation}
    \mathcal{L}_\text{attn} = \lambda_{\text{attn}} \sum_{l=1}^{5} 
    \mathrm{KL}\!\Bigl(\!\frac{1}{H}\sum_{h=1}^H A^{T}_{l,h}\,\Big\Vert\,\frac{1}{H}\sum_{h=1}^H A^{S}_{l,h}\Bigr)
\end{equation}

Here, \( T(x; W_T) \) and \( S(x'; \widetilde{W}_\text{S}(t)) \) denote the unnormalized outputs of the teacher and student ViTs respectively, where \( W_T \) are the teacher's full precision parameters and \( \widetilde{W}_\text{S}(t) \) represents the student model's progressively quantized parameters at training step \( t \). Specifically, \( \mathcal{L}_\text{cls} \) is the L2 loss between the classification tokens of the teacher and student models, $\mathcal{L}_\text{tok}$ aligns the patch token representations, and $\mathcal{L}_\text{attn}$ enforces similarity in the mean attention maps ($A_{l,h}$) across all heads $h$ in the last five layers $l$ via the KL divergence.

The overall pretraining distillation loss is then given by:

\begin{equation}
    \mathcal{L}_\text{pretrain} = \mathcal{L}_\text{cls} + \mathcal{L}_\text{tok} + \mathcal{L}_\text{attn}
\end{equation}

\subsubsection{Augmentation Strategy} As illustrated in Fig.~\ref{fig:tetra}, during pre-training we exclusively apply place-specific augmentations to generate the student input image $x'$ from the original image $x$, thereby training the student model to be invariant to the applied augmentations. Specifically, we introduce lighting variations, apply Gaussian filtering to simulate image blur, perform cropping to mimic viewpoint shifts, use ColorJitter to encourage a shape bias, and employ RandomErasing to replicate occlusions. By promoting invariance to these transformations, we enhance the model’s ability to handle diverse visual conditions.

\subsection{Fine-tuning}
After pre-training the ViT backbone with weakly-supervised learning, we fine-tune TeTRA’s local representations with an aggregation method for the VPR task using supervised contrastive learning on the GSV-Cities dataset \cite{gsvcities}. In this fine-tuning stage as shown in Fig. \ref{fig:tetra}, the ViT backbone (except for the final layer) is frozen while various aggregation modules (e.g. MixVPR, SALAD, and BoQ \cite{mixvpr, salad, boq}) are trained on-top of the backbones features. The network is fine-tuned using the multi-similarity loss \cite{multisimloss}.

To efficiently produce binary embeddings, we adopt the same progressive quantization-aware training (QAT) approach used in pre-training but replace the ternary quantization function with a binary one. This enables the model to generate binary embeddings, facilitating fast similarity search via Hamming distance, which is computed using XOR and bitcount operations \cite{bnnsurvey}. This approach significantly reduces both computational and memory demands \cite{binary}. The symmetric binary quantization function is defined as:

\begin{equation}\label{eq:sgn}
    \operatorname{sgn}(\hat{y}) =
    \begin{cases}
        -1, & \text{if } \hat{y} \le 0 \\[1mm]
         1, & \text{if } \hat{y} > 0
    \end{cases}
\end{equation}

and the progressive QAT loss is given by:

\begin{equation}\label{eq:agg_loss}
\begin{split}
\mathcal{L}_{\mathrm{agg}}(t) &= (1-\lambda(t))\mathcal{L}_{\mathrm{mult}}(\hat{y},Y) \\
&\quad + \lambda(t)\mathcal{L}_{\mathrm{mult}}(\operatorname{sgn}(\hat{y}),Y)
\end{split}
\end{equation}

Here, \(\mathcal{L}_{\mathrm{mult}}\) denotes the multi-similarity loss, \(\hat{y}\) is the network's continuous output, and \(Y\) are the target labels. The STE is used to approximate gradients for the non-differentiable binarization function.

\subsection{Implementation Details}
\textit{TeTRA} was pre-trained on the San Francisco and GSV-Cities dataset for 18 epochs using a cosine learning rate schedule with an initial rate of $1\times10^{-4}$ and a weight decay of 0.05. We used a batch size of 48 across 4 H100 GPUs with gradient accumulation (effective batch size: 384). Fine-tuning was performed on the GSV-Cities dataset for 40 epochs with a starting learning rate of $4\times10^{-4}$. A linear warm up for 3 epochs was followed by step decay (multiplier 0.3) at epochs 10, 20, and 30. We used 200 places with 4 images per place (effective batch size: 800).

\begin{figure}[h]
\centering
\includegraphics[width=\columnwidth]{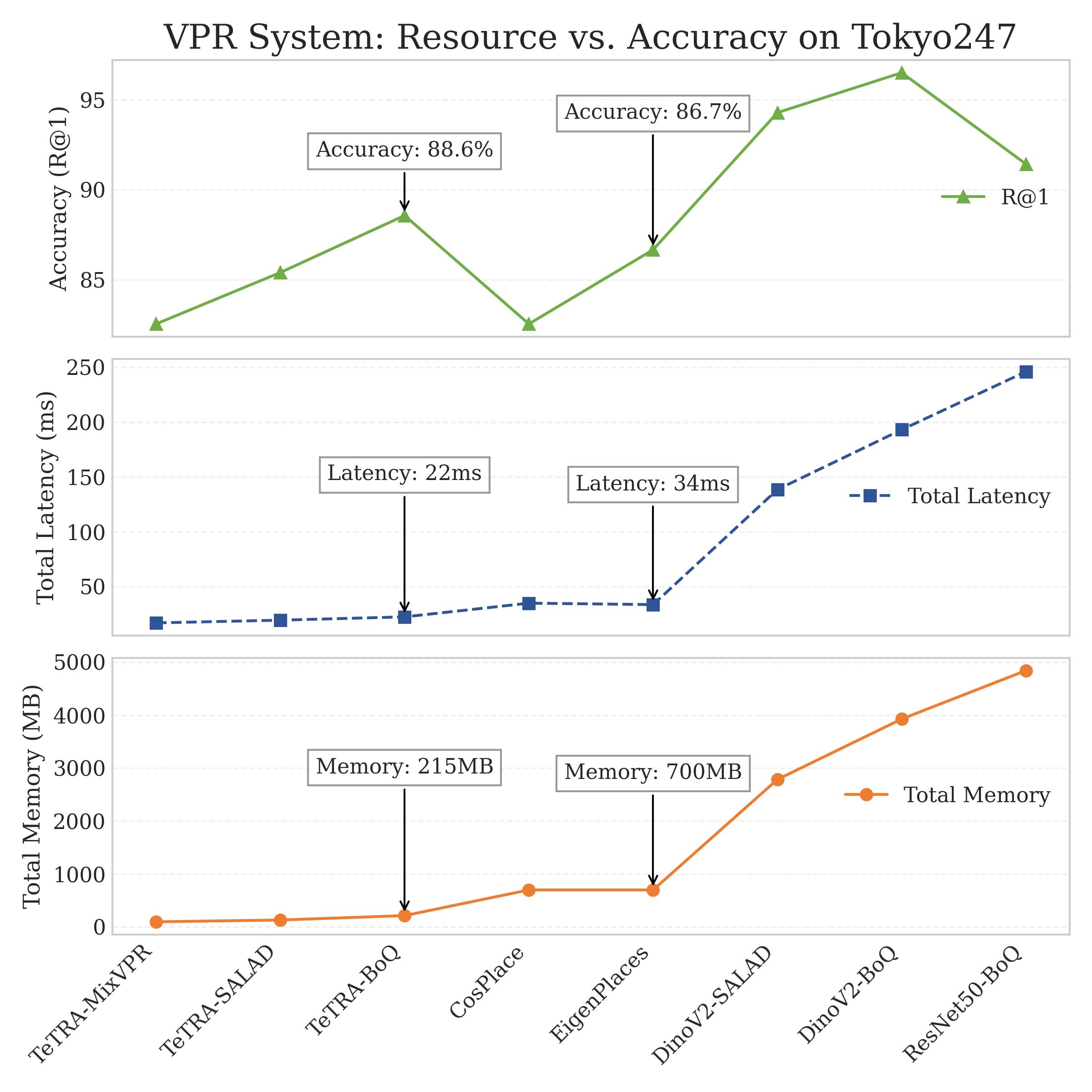}
\caption{Trade-offs in the VPR pipeline. This line plot illustrates Recall@1 accuracy, memory consumption, and latency, capturing the total computational overhead of feature extraction and matching. The figure highlights the balance between achieving high recognition performance and managing resources.}
\label{fig:fig1}
\end{figure}

\begin{table*}[!t]
  \centering
  \caption{Comparison of R@1 Accuracy and Memory Efficiency Across Datasets: Evaluating Trade-offs Between Recognition Performance and Memory Consumption}\label{tab:tab1}
  \resizebox{\textwidth}{!}{%
  \begin{tabular}{lccccccc|ccccccc}
    \toprule
    \multirow{2}{*}{Method} & \multicolumn{7}{c|}{\textbf{R@1 Accuracy (\%)}} & \multicolumn{7}{c}{\textbf{Memory Efficiency (\%/MB)}} \\
    \cmidrule(lr){2-8} \cmidrule(lr){9-15}
    & MSLS & Pitts30k & Tokyo247 & SVOX-N & SVOX-R & SVOX-S & SVOX-Sun 
    & MSLS & Pitts30k & Tokyo247 & SVOX-N & SVOX-R & SVOX-S & SVOX-Sun\\
    \midrule
    DinoV2-SALAD \cite{salad} & 88.2 & 92.4 & 94.3 & 95.5 & 98.7 & 98.6 & 97.1 & 0.1 & 0.1 & 0.0 & 0.1 & 0.1 & 0.1 & 0.1 \\
    DinoV2-BoQ \cite{boq} & \textbf{89.9} & \textbf{93.7} & \textbf{96.5} & \textbf{97.2} & \textbf{98.8} & \textbf{99.4} & \textbf{97.5} & 0.1 & 0.1 & 0.0 & 0.1 & 0.1 & 0.1 & 0.1 \\
    ResNet50-BoQ \cite{boq} & 87.7 & 92.4 & 91.4 & 87.1 & 96.2 & 98.7 & 95.8 & 0.1 & 0.1 & 0.0 & 0.1 & 0.1 & 0.1 & 0.1 \\
    EigenPlaces-D256 \cite{eigenplaces} & 83.1 & 91.3 & 78.7 & 42.5 & 85.3 & 90.7 & 76.1 & 0.8 & 0.9 & 0.5 & 0.4 & 0.8 & 0.8 & 0.7 \\
    EigenPlaces-D512 \cite{eigenplaces} & 84.9 & 92.1 & 83.5 & 53.7 & 88.5 & 91.0 & 79.9 & 0.6 & 0.8 & 0.3 & 0.4 & 0.7 & 0.7 & 0.6 \\
    EigenPlaces-D2048 \cite{eigenplaces} & 85.6 & 92.7 & 86.7 & 59.0 & 88.2 & 91.4 & 86.1 & 0.3 & 0.5 & 0.1 & 0.2 & 0.4 & 0.4 & 0.4 \\
    CosPlaces-D256 \cite{cosplace} & 82.9 & 90.4 & 82.5 & 46.0 & 82.1 & 89.1 & 69.7 & 0.8 & 0.9 & 0.5 & 0.4 & 0.8 & 0.8 & 0.6 \\
    CosPlaces-D512 \cite{cosplace} & 83.3 & 90.2 & 84.4 & 49.9 & 84.7 & 87.9 & 72.8 & 0.6 & 0.8 & 0.3 & 0.4 & 0.7 & 0.7 & 0.6 \\
    CosPlaces-D2048 \cite{cosplace} & 84.1 & 90.8 & 82.5 & 49.6 & 85.4 & 90.2 & 76.3 & 0.3 & 0.5 & 0.1 & 0.2 & 0.4 & 0.4 & 0.3 \\
    \midrule
    FloppyNet \cite{binary}& 22.2 & 46.1 & 0.2 & 0.0 & 0.0 & 0.1 & 0.1 & 0.0 & 0.0 & 0.1 & 0.0 & 0.0 & 0.0 & 0.0 \\
    NetVLADSP \cite{pruning} & 78.6 & 89.1 & 67.6 & 46.4 & 85.4 & 91.9 & 76.6 & 0.1 & 0.2 & 0.3 & 0.1 & 0.2 & 0.2 & 0.2  \\
    DinoV2-SALAD-INT8 \cite{salad} & 62.9 & 82.6 & 56.8 & 10.1 & 45.8 & 58.4 & 30.9 & 0.1 & 0.2 & 0.0 & 0.0 & 0.1 & 0.1 & 0.0 \\
    DinoV2-BoQ-INT8 \cite{boq} & 62.6 & 81.5 & 54.6 & 4.5 & 37.4 & 47.4 & 26.6 & 0.1 & 0.1 & 0.0 & 0.0 & 0.0 & 0.1 & 0.0 \\
    ResNet50-BoQ-INT8 \cite{boq} & 87.0 & 92.1 & 90.8 & 77.8 & 95.4 & 95.2 & 93.6 & 0.1 & 0.1 & 0.0 & 0.1 & 0.1 & 0.1 & 0.1 \\
    EigenPlaces-D2048-INT8 \cite{eigenplaces} & 84.4 & 91.5 & 87.3 & 49.8 & 85.0 & 89.3 & 75.5 & 0.5 & 0.9 & 0.1 & 0.3 & 0.5 & 0.6 & 0.5 \\
    CosPlaces-D2048-INT8 \cite{cosplace} & 83.4 & 86.9 & 87.3 & 43.1 & 84.0 & 87.5 & 69.6 & 0.5 & 0.8 & 0.1 & 0.3 & 0.5 & 0.5 & 0.4 \\
    \midrule
    \textit{TeTRA}-BoQ & 86.6 & 91.7 & 88.6 & 64.9 & 96.7 & 98.0 & 93.9 & 0.7 & 0.8 & 0.4 & 0.5 & 0.8 & 0.8 & 0.7 \\
    \textit{TeTRA}-SALAD & 84.7 & 90.3 & 85.4 & 52.4 & 94.5 & 96.3 & 93.4 & 1.1 & \textbf{1.4} & 0.6 & \textbf{0.7} & \textbf{1.3} & 1.3 & \textbf{1.3} \\
    \textit{TeTRA}-MixVPR & 82.2 & 90.3 & 82.5 & 49.0 & 92.2 & 94.9 & 90.8 & \textbf{1.2} & \textbf{1.4} & \textbf{0.8} & \textbf{0.7} & \textbf{1.3} & \textbf{1.4} & \textbf{1.3} \\
    \bottomrule
  \end{tabular}%
  }

\end{table*}\label{tab:tab1}

\section{Results}\label{sec:results}
We organize our evaluation into three main subsections. First, we analyze the matching phase alone (Sec.~\ref{sec:matching_perf}). Next, we consider the entire VPR pipeline, including feature extraction and matching (Sec.~\ref{sec:whole_perf}). Finally, we present a comprehensive analysis of robustness to appearance changes (Sec.~\ref{sec:appearance}), followed by conclusions.

\subsection{Matching Performance}\label{sec:matching_perf}
This subsection focuses on the matching phase and compares the performance of efficient baseline approaches. Fig. \ref{fig:matching} compares the resource and accuracy trade-offs of EigenPlaces and CosPlace models across descriptor sizes ranging from 2048 to 32 in terms of database memory and matching latency. In these models the descriptor size variations reveal clear trade-offs between memory requirements and latency. We exclude DinoV2 based methods as they exhibit significantly higher resource consumption (see Fig. \ref{fig:fig1}) and are thus outliers in Fig. \ref{fig:matching}.

In contrast, \textit{TeTRA} employs 1-bit embeddings, enabling efficient Hamming distance calculations via XOR and bitcount instructions~\cite{binary}. As shown in Fig.~\ref{fig:matching}, these binary embeddings significantly reduce matching latency, even for larger descriptor dimensions. \textit{TeTRA} surpasses baseline trends by providing a more favorable trade-off between memory usage and latency, thereby delivering a Pareto-optimal solution across all descriptor sizes. This performance advantage holds for all aggregation variants of \textit{TeTRA}, including BoQ, MixVPR, and SALAD, showcasing its improved resource and accuracy trade-off in matching performance.

\subsection{Performance of the Entire VPR Pipeline}\label{sec:whole_perf}
We next evaluate the trade-offs between resource consumption and accuracy within the complete VPR pipeline, which encompasses both feature extraction and matching. Fig.~\ref{fig:fig1} presents Recall@1 accuracy alongside memory usage and latency, measured as the cumulative cost of these two stages. As illustrated, transformer-based methods, particularly those leveraging DinoV2, exhibit significantly higher memory and latency demands. This trend is also observed in ResNet-BoQ, which incurs substantial memory overhead due to its 16,384-dimensional descriptor. In contrast, EigenPlaces and Cosplace demonstrate superior resource efficiency, benefiting from compact descriptors and lightweight convolutional architectures. Notably, \textit{TeTRA}, despite employing a transformer backbone, further enhances efficiency, reducing memory consumption by $69\%$ and latency by $35\%$ relative to EigenPlaces. Crucially, despite these reductions in resource usage, \textit{TeTRA}-BoQ surpasses EigenPlaces and Cosplace in Recall@1 accuracy, underscoring its effectiveness in VPR applications.

\subsection{Memory Efficiency and Robustness}\label{sec:appearance}
Table \ref{tab:tab1} compares the R@1 and memory efficiency scores of \textit{TeTRA} variants and various baseline models across multiple benchmark datasets. These include Mapillary Street Level Sequences (MSLS), Pittsburgh30k (Pitts30k), Tokyo247, and SVOX appearance-change datasets, where SVOX-Night is denoted as SVOX-N, SVOX-Rain as SVOX-R, SVOX-Snow as SVOX-S, and SVOX-Sun as SVOX-Sun. The table shows that under large appearance changes, \textit{TeTRA} consistently achieves higher Recall@1 compared to efficient convolutional models (EigenPlaces, CosPlace) while significantly reducing memory requirements. Even with aggressive 2-bit ternary quantization of the ViT backbone and 1-bit binarization of embeddings, its performance remains close to that of full-precision Transformers such as DinoV2-BoQ. Notably, on the SVOX-night dataset, \textit{TeTRA}-BoQ yields higher Recall@1 than CosPlace and EigenPlaces by 15.3 and 5.9 percent respectively, underscoring its robustness to lighting variations. Furthermore, this trend of improved accuracy also holds for the DinoV2 INT8 quantized baselines, given that post-training quantization on ViTs is challenging due to activation outliers. Additionally, although \textit{TeTRA} uses only a single extra bit per weight compared to FloppyNet, a fully binarized approach, it performs significantly better, particularly on SVOX appearance change datasets where FloppyNet fails. Finally, when compared to the structured pruning approach used by NetVLADSP~\cite{pruning}, \textit{TeTRA} achieves higher recall@1 on every dataset.

The last seven columns of Table~\ref{tab:tab1} report memory efficiency scores (Recall@1/MB), where \textit{TeTRA} consistently attains higher values than all baselines. For instance, on the  SVOX-Sun dataset, \textit{TeTRA}-SALAD and \textit{TeTRA}-MixVPR's memory efficiency scores exceed 1.3\%/MB, more than double the memory efficiency of the other baselines, reflecting their strong trade-off between accuracy and model size. 

The radar plots in Fig.~\ref{fig:radar} further illustrate this balance, showing that full-precision ViTs (DinoV2-BoQ and DinoV2-SALAD) cluster in the high-accuracy region but require significantly more memory. In contrast, \textit{TeTRA} remains near the outer edge of the memory efficiency axis while staying highly competitive with full-precision DinoV2-based approaches in R@1 and outperforming CosPlace, a convolution-based model, particularly on appearance-change datasets. This advantage is critical for deployment on resource limited platforms such as drones or mobile robots, where on-board storage and compute are constrained. Overall, these results underscore that carefully designed extreme low-bit quantization, combined with distillation and progressive QAT, can achieve high Recall@1 accuracy in diverse conditions, even when comparing to the most efficient convolutional models. 

\section{Conclusion}\label{sec:conclusion}
In this work, we introduced \textit{TeTRA} as a ternary transformer approach for visual place recognition that substantially reduces memory consumption while preserving retrieval accuracy. By quantizing the Vision Transformer backbone to two bits and binarizing the final embeddings with a carefully structured distillation process, \textit{TeTRA} achieves up to 69 percent lower memory usage and 35 percent lower latency compared to leading efficient convolutional architectures. This approach incurs a negligible loss in retrieval accuracy, making \textit{TeTRA} a state-of-the-art solution for resource-constrained applications. Future work includes extending this method to sequential based systems and adding an efficient two-stage retrieval mechanism to further boost accuracy.

\bibliographystyle{IEEEtran}
\bibliography{references}

\newpage

\vfill

\end{document}